# A Novel Neural Network Training Framework with Data Assimilation

Chong Chen[1, *], Qinghui Xing[1], Xin Ding[1], Yaru Xue[1], Tianfu Zhong[1]

[1] College of Information Science and Engineering, China University of Petroleum – Beijing

[*] Email: chenchong@cup.edu.cn

**Abstract**

In recent years, the prosperity of deep learning has revolutionized the Artificial Neural Networks. However, the dependence of gradients and the offline training mechanism in the learning algorithms prevents the ANN for further improvement. In this study, a gradient-free training framework based on data assimilation is proposed to avoid the calculation of gradients. In data assimilation algorithms, the error covariance between the forecasts and observations is used to optimize the parameters. Feedforward Neural Networks (FNNs) are trained by gradient decent, data assimilation algorithms (Ensemble Kalman Filter (EnKF) and Ensemble Smoother with Multiple Data Assimilation (ESMDA)), respectively. ESMDA trains FNN with pre-defined iterations by updating the parameters using all the available observations which can be regard as offline learning. EnKF optimize FNN when new observation available by updating parameters which can be regard as online learning. Two synthetic cases with the regression of a Sine Function and a Mexican Hat function are assumed to validate the effectiveness of the proposed framework. The Root Mean Square Error (RMSE) and coefficient of determination ($R^2$) are used as criteria to assess the performance of different methods. The results show that the proposed training framework performed better than the gradient decent method. The proposed framework provides alternatives for online/offline training the existing ANNs (e.g., Convolutional Neural Networks, Recurrent Neural Networks) without the dependence of gradients.

Keywords—Neural network, Training algorithm, data assimilation, EnKF, ESMDA

## 1. Introduction

Artificial Neural Networks (ANNs) have been investigated and utilized extensively by researchers in numerous fields to conduct predictions and classifications based on the knowledge learning from training data (LeCun, Bengio et al. 2015, Huang, Gao et al. 2019). Significant accomplishments have been achieved by applying ANNs in computer vision, speech recognition and natural language processing (Jin, McCann et al. 2017, Zhou, Chen et al. 2019). ANN was first inspired by the biological neural networks which constitute animal brains (McCulloch and Pitts 1943). ANN was a mathematical model of biological neural networks with neurons, connections (axons) and transfer functions (synapse).

After decades of researches and developments, ANNs have evolved from Perceptron (Rosenblatt 1957) to Hopfield network (Hopfield 1982), to Back Propagation Neural Network (Rumelhart, Hinton et al. 1986) and more recently to deep learning (LeCun, Bengio et al. 2015) which makes ANNs one of the most important Feedforward Neural Networks (FNNs). Non-linear mapping capability was obtained by applying sufficiently large number of neurons, connections, weights, bias, transfer functions and learning algorithms. ANNs are capable of approximate any function with any given precision from a mathematical perspective (Cybenko 1989, Hornik 1991). However, critical issues should be addressed for applying ANN more effectively.

One of the most important issue is the dependence of gradient during training ANNs. The number of neurons, connections, weights, bias, transfer functions is essential aspects should be considered while constructing ANNs. A training procedure which adjusts the weights and bias is necessary to ensure the behavior of ANNs as expected. Backpropagation has played an important role since 1980s which is efficient for training ANNs with a teacher-based supervised learning algorithm. The errors are backpropagated through the networks based on gradient decent algorithm. The algorithm might be trapped in local minima because of the dependence of local gradient information. Although some improved methods (e.g., Batch Gradient Descent, Stochastic Gradient Descent and Mini-batch Gradient Descent) have been proposed, the convergence of ANNs during training stages is another problem which would further influence the performance of training and predicting. Therefore, some researchers tend to train ANNs with Heuristic Algorithm (HA). Professor Zhao Hong proposed General Vector Machine (GVM) which trains ANNs with Monte Carlo algorithm and Least Squares

(Zhao 2016). The generalization and prediction ability performed well in relatively small data sets, but this method could hardly obtain satisfied results in large data sets with the increase of computation cost exponentially. Simulated Annealing was integrated with Gradient Descent and Backpropagation to avoid local optima during training ANNs (Khan, Hameed et al. 2019). Researches and progresses have been acquired by applying Genetic Algorithm to adjust the weights and bias during training procedure. However, solid theoretical basis was missing due to the origination of HAs.

Data Assimilation (DA) is originated from and has a long tradition in meteorology and oceanography (Daley 1993, Houtekamer and Zhang 2016). The essence of DA is to deal with uncertainty by assimilate different kinds of observations. It is well known that a free-running model will accumulate errors until its prediction is no long useful (Tribbia and Baumhefner 2004). The only way to avoid this procedure is to allow the model influenced by observations (Leith 1993). DA provide a solution to evolve the models by involving available observations. Different names are used in different fields, e.g. state estimation (Wunsch 2006); optimization (Biegler, Coleman et al. 2012); history matching (Emerick 2012); retrieval production (Rodgers 2000); inverse modeling (Tarantola 2005). The objective of DA is to produce information about the posterior probability density function (PDF) by different approaches. There are three categories of Bayesian-based strategies of DA methods: (1) Variational DA with implementations of 3D-Var or 4D-Var; (2) Ensemble DA which implements based on Ensemble Kalman Filter (EnKF); (3) Monte-Carlo methods which allow the assimilation of information with non-Gaussian errors. The EnKF (Evensen 1994) derived from the merge of Kalman Filter (Kalman 1960) and Monte Carlo estimation methods (Kalman and Bucy 1961). The algorithm has been examined and applied in various fields such as metrology, oceanography, petroleum engineering and hydrogeology (Hendricks Franssen and Kinzelbach 2008, Aanonsen, Nævdal et al. 2009, Erazo, Wallscheid et al. 2020), since it was first proposed by Evensen (Evensen 1994). The simple conceptual formulation and relative ease of implementation (no derivation of a tangent linear operator or adjoint equations are required) with affordable computational requirements results in the popularity of EnKF. The system states and parameters can be forecasted and updated simultaneously with minimized error covariance. Ensemble Smoother with Multiple Data Assimilation (ESMDA) (Emerick and Reynolds 2013) was then introduced based on the Ensemble Smoother (ES) proposed by van Leeuwen and Evensen (Leeuwen and Evensen 1996) in order to avoid stopping and restart the model run when observations happen. A range of methods based on Monte Carlo techniques are formed to conduct DA. The Particle Filter (PF) represents a PDF by ensembles (particles) without the limitation of Gaussianity of the distribution. DA algorithms offer an opportunity for optimizing the parameters, quantifying the uncertainty and gradient-free training of ANNs at the same time.

In this paper, a novel training framework for ANNs was proposed by adopting data assimilation to avoid the dependence of gradient and hence some disadvantages of gradient-descent-based methods. To illustrate the idea, a fully connected FNN integrated with EnKF and ESMDA were implemented. Two synthetic cases with the regression problem of Sine function and Mexican Hat function were conducted to test and validated the proposed framework. The paper is organized as follows. Section 2 provides the theory of FNN, data assimilation and the proposed framework. Section 3 presents the data and setting of a synthetic case to validate the proposed framework. The results are demonstrated in Section 4. Finally, a summary and conclusions are given in Section 5.

## 2. Methodology

### 2.1. Feedforward Neural Network

A Feedforward Neural Network (FNN) is an ANN wherein the information flows from the input layer through the transfer functions to the output layer. There are no feedback connections and hence the nodes (neurons) do not form a cycle. Neurons were proposed by Frank Rosenblatt (Rosenblatt 1957) inspired by Warren McCulloch and Walter Pitts (McCulloch and Pitts 1943). In a neuron, the output is calculated by a nonlinear function (activation function or transfer function) of the sum of its inputs as $y = \sum_{i=1}^{p} f(w_i i_i)$. An FNN is formed by the combination of neurons as in the biological neural networks. Because of the typicality and comprehensibility, the three-layer FNN (input layer, hidden layer and output layer) of neurons is used in this study (Fig. 1). The feedforward process is the same as common fully-connected neural networks as follows:

$$h_j = f_1\left(\sum_{i=1}^{n} w_{ji} \times x_i\right) \quad i = 1,2,\dots,n; j = 1,2,\dots,N_h \tag{1}$$

$$y_k = f_2\left(\sum_{j=1}^{N_h} w_{kj} \times h_j + b_j\right) \quad k = 1,2,\dots,m \tag{2}$$

where $x_i$, $h_j$ and $y_k$ represent the nodal values in the input layer, hidden layer and output layer,

respectively; $n$, $N_h$ and $m$ are the number of neurons in the input layer, hidden layer and output layer; $w_{ji}$ is the weight connecting the input $x_i$ and the $j$th neuron in the hidden layer; $b_j$ represents the bias in the output layer; $w_{kj}$ is the weight connecting the $j$th neuron in the hidden layer ($h_j$) and the output $y_k$; $f_1$ and $f_2$ are the activation functions in the hidden layer and the output layer.

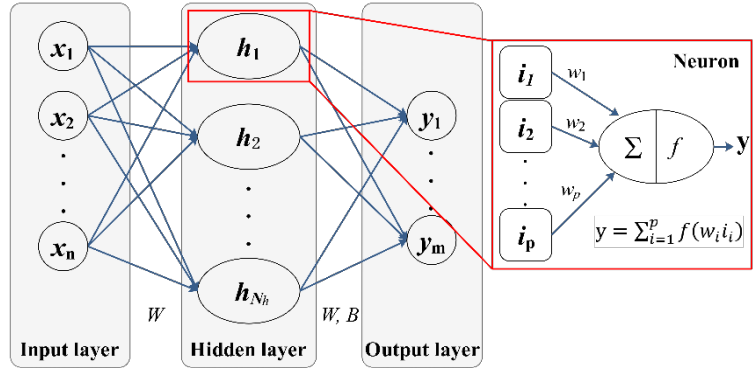

**Fig 1.** The structure of Feedforward Neural Networks and neurons

## 2.2. Data assimilation

Generally, data assimilation combines information from a variety of sources to improve the accuracy of predictions and takes the uncertainty from measurements, inputs, parameters and model structures into account at the same time. In a nonlinear dynamic system, the state vector which contains both the states and the parameters is defined as follows:

$$X_t = \begin{pmatrix} A \\ B \end{pmatrix} \quad t = 1, 2, \dots, N_t \tag{3}$$

where $X_t$ is the state vector at time $t$ with the dimension of $N_x \times N_t$; $N_t$ denotes the number of time steps; $A$ represents the parameters vector with dimension of $N_a \times N_t$; $B$ represents the states with dimension of $N_b \times N_t$; $N_x$ denotes the number of state variables in $X_t$ which equals $N_a + N_b$.

The system is treated as derivations of state equation (Eq. 4) and observation equation (Eq. 5) through time $t$.

$$X_t^f = M_{t-1}(X_{t-1}^a) + \xi \qquad \xi \sim N(0, \Xi) \tag{4}$$

$$Y_t = H X_t^f \tag{5}$$

where $f$ denotes the forecast (prior estimation) of the parameters and states; $a$ denotes the analysis (posterior estimation) of the parameters and states; $X_t^f$ represents the forecast of the parameters and states at time $t$; $M_{t-1}$ is the nonlinear model operator; $X_{t-1}^a$ is the analysis of the parameters and states at time $t$-1; $\xi \sim N(0, \Xi)$ indicates a Gaussian distribution with zero mean and covariance matrix $\Xi$;

$Y_t$ is the observation vector at time $t$; $H$ represents the observation operator which connects the model parameters and the observations.

**2.3. Ensemble Kalman Filter**

EnKF algorithm necessarily includes forecast and analysis steps.

In the forecast step, the forecasted parameters and states is updated according to Eq. (4).

In the analysis step, the observation data are first perturbed by random errors:

$$Y_t^o = Y^o + \varepsilon \qquad \varepsilon \sim N(0, R_e) \qquad (6)$$

where $Y^o$ represents the observation data at time $t$; $\varepsilon \sim N(0, R_e)$ indicates Gaussian random observation errors with zero mean and covariance matrix $R_e$.

The analysis of states and parameters are obtained by updating the forecast as follows:

$$X_t^a = X_t^f + \frac{P_t^f H^T}{H P_t^f H^T + R_e}(Y_t^o - Y_t) \qquad (7)$$

$$P_t^a = \left(I - \frac{P_t^f H^T}{H P_t^f H^T + R_e} H\right) P_t^f \qquad (8)$$

Here

$$P_t^f = \frac{1}{N_e - 1} \sum_{i=1}^{N_e} \left(x_{i,t}^f - \overline{x_t^f}\right)\left(x_{i,t}^f - \overline{x_t^f}\right)^T \qquad (9)$$

$$\overline{x_t^f} = \frac{1}{N_e} \sum_{i=1}^{N_e} x_{i,t}^f \qquad (10)$$

Define

$$K_t = \frac{P_t^f H^T}{H P_t^f H^T + R_e} \qquad (11)$$

Where $K_t$ is the Kalman gain matrix at time $t$; $N_e$ represents the ensemble size: $P_t^f$ is the covariance matrix of the forecast at time $t$; $\overline{x_t^f}$ is the mean of ensemble members for forecast states and parameters.

Together with Eq. (4) and Eq. (5), EnKF is able to dynamically update the system estimates when new observations become available.

**2.4. Ensemble Smoother with Multiple Data Assimilation**

The ensemble based sequential data assimilation (e.g. EnKF, PF) updates the parameters and states at the time when observations happen which requires to restart the simulations. The recurrent simulation may be inconvenient when the purpose is to incorporate different kinds of data for history matching. Therefore, Ensemble Smoother with Multiple Data Assimilation (ESMDA) is proposed to obtain better data matches and lower computation cost. Unlike sequential data assimilation, ESMDA computes a global update by simultaneously assimilating all the available data several times.

ESMDA is an iterative Ensemble Smoother with a predefined number of iterations for data assimilation. An inflation coefficient $\alpha_i$ is introduced to the measurement error in each iteration. The requirement of inflation coefficient is described in Eq. (12) to maintain correct posterior mean and covariance for linear cases.

$$\sum_{i=1}^{N_i} \frac{1}{\alpha_i} = 1 \qquad (12)$$

where $N_i$ is the predefined number of iterations for data assimilation. Apparently, there are many alternatives for inflation coefficient which satisfies the requirement. The determination of $\alpha_i$ refer to (Emerick and Reynolds 2012).

The inflation coefficient is used to inflate the perturbation of all observation data and its covariance matrix in Eq. (13) and Eq. (14) which leads to:

$$Y = Y^o + \sqrt{\alpha_i}\varepsilon \qquad \varepsilon \sim N(0, R_e) \qquad (13)$$

$$K = \frac{P^f H^T}{H P^f H^T + \sqrt{\alpha_i} R_e} \qquad (14)$$

### 2.5. Training FNN with DA

Assume the structure (the number of layers, the nodes in each layer and the connection between nodes) of FNN for a specified problem is determined and represented by $M^*$. The weights ($w$ in Eq. (1) and (2)) and bias ($b$ in Eq. (2)) are regarded as states ($X^*$) of $M^*$ which leads to $X^* = \binom{w}{b}$.

In the perspective of DA, substitute $M$ in Eq. (4) with $M^*$, we can obtain:

$$X_t^{*f} = M_{t-1}^*(X_{t-1}^{*a}) + \xi^* \qquad \xi \sim N(0, \Xi) \qquad (15)$$

$$Y_t^* = H^*(X_t^{*f}) \qquad (16)$$

Where $Y_t^*$ represents the outputs of $M^*$ with the element of $y_k$ in Eq. (2); $X^*$ is the parameters which can be updated by Eq. (7) to (11).

In the perspective of FNN, the optimization of parameters (*w* and *b*) in the back propagation process is replaced by data assimilation. The ESMDA can be used to train the FNN with the historical data. The sequential data assimilation can be used to adjust the model trained by ESMDA with the real-time observations. The procedure of FNN trained by sequential data assimilation and ESMDA is shown in Fig. 2.

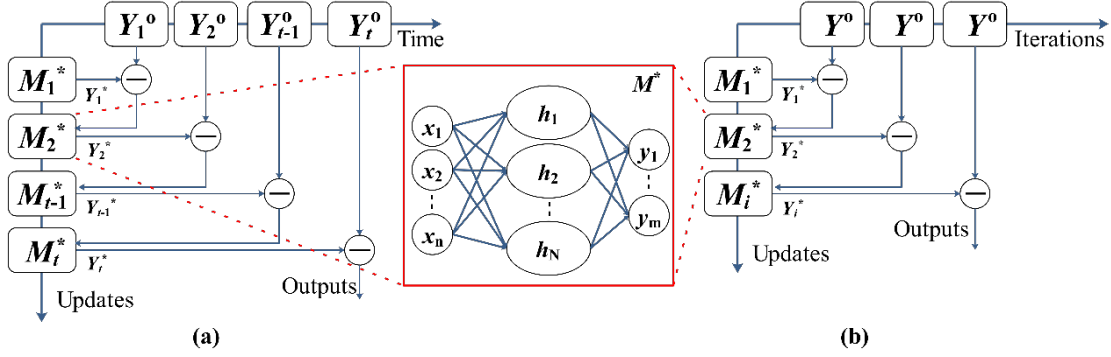

**Fig. 2.** The procedure of FNN trained by (a) sequential data assimilation; (b) ESMDA.

The combination of FNN and DA can be summarized as Algorithm 1 (for EnKF) and Algorithm 2 (for ESMDA). There are several hyper-parameters for Algorithm 1 and Algorithm 2 which should be determined based on the prior information of the actual situation.

---

**Algorithm 1.** Training FNN with EnKF

**Input:** *x*; **Output:** *y*

**Trainable parameters:** *w* and *b*

**Hyper-parameters:** $N_h$, $N_e$, $\Xi$, $R_e$

---

Construct $M^*$;

Generate initial parameter ensembles $X_0^f$ with $N(0, \Xi)$

for $t = 1$ to $N_t$

    for *ens_num* = 1 to $N_e$

        $h_j = f_1\left(\sum_{i=1}^{n} w_{ji} \times x_i\right)$     $i = 1, 2, \ldots, n;\; j = 1, 2, \ldots, N_h$

        $y_k = f_2\left(\sum_{j=1}^{N_h} w_{kj} \times h_j + b_j\right)$     $k = 1, 2, \ldots, m$

    endfor

    $Y_t = y$

    Generate perturbed observations $Y_t^o$ with $N(0, R_e)$

    $\overline{x_t^f} = \frac{1}{N_e} \sum_{i=1}^{N_e} x_{i,t}^f$

$$P_t^f = \frac{1}{N_e-1}\sum_{i=1}^{N_e}\left(x_{i,t}^f - \overline{x_t^f}\right)\left(x_{i,t}^f - \overline{x_t^f}\right)^T$$

$$X_t^a = X_t^f + \frac{P_t^f H^T}{HP_t^f + R_e}(Y_t^o - Y_t)$$

$$P_t^a = \left(I - \frac{P_t^f H^T}{HP_t^f H^T + R_e}H\right)P_t^f$$

$$X_{t+1}^f = X_t^a$$

endfor

---

**Algorithm 2.** Training FNN with ESMDA

---

**Input:** *x*; **Output:** *y*

**Trainable parameters:** *w* and *b*

**Hyper-parameters:** $N_h$, $N_i$, $N_e$, $\Xi$, $R_e$

---

Construct $M^*$;

Generate initial parameter ensembles $X_0^f$ with $N(0, \Xi)$

for *iteration* = 1 to $N_i$

    for *ens_num* = 1 to $N_e$

$$h_j = f_1\left(\sum_{i=1}^{n} w_{ji} \times x_i\right) \quad i = 1,2,\ldots,n; \; j = 1,2,\ldots,N_h$$

$$y_k = f_2\left(\sum_{j=1}^{N_h} w_{kj} \times h_j + b_j\right) \quad k = 1,2,\ldots,m$$

    endfor

    $Y_{iteration} = y$

    Generate perturbed observations $Y^o_{iteration}$ with $N(0, R_e)$

$$\overline{x_{iteration}^f} = \frac{1}{N_e}\sum_{i=1}^{N_e} x_{i,iteration}^f$$

$$P_{iteration}^f = \frac{1}{N_e-1}\sum_{i=1}^{N_e}\left(x_{i,iteration}^f - \overline{x_{iteration}^f}\right)\left(x_{i,iteration}^f - \overline{x_{iteration}^f}\right)^T$$

$$X_{iteration}^a = X_{iteration}^f + \frac{P_{iteration}^f H^T}{HP_{iteration}^f + \sqrt{\alpha_k}R_e}(Y_{iteration}^o - Y_{iteration})$$

$$P_{iteration}^a = \left(I - \frac{P_{iteration}^f H^T}{HP_{iteration}^f H^T + \sqrt{\alpha_k}R_e}H\right)P_{iteration}^f$$

$$X_{iteration+1}^f = X_{iteration}^a$$

endfor

---

## 3. Synthetic cases

The performance of the proposed integration of FNN and DA was validated through two synthetic cases. The main purposes of the synthetic cases are: (1) to understand the proposed method; (2) to analyze the capability of the proposed method in generating accurate estimations without gradient information by comparing the performance of the proposed method with the traditional gradient decent method. In the synthetic cases, a one-dimensional regression dataset which generated from Sine function and Mexican Hat function were utilized. Different methods were conducted and developed to optimize the FNN model. The methods used in the synthetic cases were summarized in Table 1.

**Table 1.** Methods used in synthetic cases.

| Datasets | Model | Optimization of FNN | Performance criteria |
|---|---|---|---|
| Sine function | FNN | EnKF    ESMDA | RMSE    $R^2$ |
| Mexican Hat function | FNN | EnKF    ESMDA | RMSE    $R^2$ |

### 3.1. Performance criteria

As recommended by (Moriasi 2007), the Root Mean Square Error (RMSE) and Coefficient of Determination ($R^2$) were used as objective functions to assess the performance of the two synthetic cases (as shown in Equation (20) and (21)). The RMSE measures the average magnitude of the error between model simulations (M) and observations (O). As shown in Equation (20), the errors are squared before averaged, large errors take a relatively high weight. Therefore, RMSE is useful when large errors are undesirable and $R^2$ measures the predictive ability of models.

$$RMSE = \sqrt{\frac{1}{N}\sum_{i=1}^{N}(M_i - O_i)^2} \quad (20)$$

$$R^2 = 1 - \frac{\sum_{i=1}^{N}(O_i - M_i)^2}{\sum_{i=1}^{N}(O_i - \bar{O})^2} \quad (21)$$

Where N represents the total number of observations; $\bar{O}$ is the average of observations.

Besides, the computation time was also recorded as a criterion to assess the computation costs of different models.

### 3.2. Data

In the Sine function case, two datasets (training data and validation data) were generated from sine function. The data in training stage was generated in (0, 2π) with interval of 0.01π which resulted in

201 samples. The data in validation stage was generated in (0, 2π) with interval of 0.1π which resulted in 21 samples. Detail information of the data was summarized in Table 2.

**Table 2.** Data description for the Sine function case

|  | Training stage | | | Validation stage | | |
| --- | --- | --- | --- | --- | --- | --- |
|  | Number of samples | Range | Interval | Number of samples | Range | Interval |
| Input | 201 | [0,2π] | 0.01π | 21 | [0,2π] | 0.1 |
| Output | 201 | [-1, 1] | * | 21 | [-1, 1] | * |

In the Mexican Hat function case, two datasets (training data and validation data) were generated using $\psi(t) = \frac{2}{\sqrt{3}\sigma\pi^{\frac{1}{4}}}\left(1-\left(\frac{t}{\sigma}\right)^2\right)e^{-\frac{t^2}{2\sigma^2}}$ with σ=1. 200 samples were generated in (-5, 5) to formulate the training dataset. The data in validation stage was generated in (-5, 5) with the number of samples being 30. Detail information of the data was summarized in Table 3.

**Table 3.** Data description for the Mexican Hat function case.

|  | Training stage | | Validation stage | |
| --- | --- | --- | --- | --- |
|  | Number of samples | Range | Number of samples | Range |
| Input | 200 | [-5, 5] | 30 | [-5, 5] |
| Output | 200 | $[-\frac{4}{\sqrt{3}\pi^{\frac{1}{4}}}e^{-\frac{3}{2}}, \frac{2}{\sqrt{3}\pi^{\frac{1}{4}}}]$ | 30 | $[-\frac{4}{\sqrt{3}\pi^{\frac{1}{4}}}e^{-\frac{3}{2}}, \frac{2}{\sqrt{3}\pi^{\frac{1}{4}}}]$ |

### 3.3. Settings

The architectures and parameters of the two synthetic cases remained identical. The architecture of FNN was predefined to be fully connected network with one input layer, one hidden layer and one output layer. Based on the features of the dataset, the number of neurons in each layer is one, ten and one. respectively. The parameters (weights and biases) were randomly initialized from a normal distribution. Without loss of generality, the biases between the hidden layer and output layer were selected to be assimilated in EnKF. Regarding the hyperparameters in the EnKF, the ensemble size $N_e$, the prior parameter covariance matrix $\Xi$ and the observation error covariance matrix $R_e$ are 50, 0.1 and 0.005, respectively. The observation covariance $R_e$ was set to be a small value because the observations used in EnKF were generated from the Sine wave which was accurate and much more trustworthy than the FNN model. The loss function in gradient decent method is Mean Square Error (MAE). 10000 epochs with learning rate of 0.12 were used for gradient decent method to train the FNN.

### 4. Results and Discussions

#### 4.1. Performance of FNN model optimized by EnKF

In the synthetic case of Sine function, the results calculated from FNN which optimized by different methods were shown in Fig. 4. After 10000 epochs of training, the FNN model with gradient decent method approaches the Sine Function with some biases. The RMSE and $R^2$ value for FNN model optimized by gradient decent were 0.0948 and 0.9819. Although the values of performance criteria were relatively acceptable, there were still some biases in the peak and trough of the wave which may because of the difference of gradient changes and static learning rate of the algorithm. In the experiment of EnKF-optimized FNN, there were ensembles for the parameters which generated from a random normal distribution. To calculate the performance criteria, the ensemble mean was used as the final model outputs. The FNN model optimized by EnKF indicated a better match to the Sine Function (the red curve in Fig. 4) with RMSE of 0.0317 and $R^2$ of 0.9980. Each realizations of parameters could be regarded as a possible realization of FNN. On the contrary to the gradient decent algorithm, EnKF was capable of capturing the variance of gradient changes because of the updating procedure in EnKF algorithm. The evolution of parameters (shown in Fig. 5) also reflected the correction processes of the parameters to adapt the larger gradient changes. After randomly generating the FNN parameters (biases from the hidden layer to the output layer), the uncertainty of parameter remains relatively large because of the large difference between the FNN model and the Sine wave according to Eq. (7). The same situation could be found at "$x = 1.5\pi$". On the contrary, the parameter uncertainty was reduced when "$x \in (0.75\pi, 1.25\pi)$" because of the relatively small difference between the FNN model and the Sine wave (Fig. 5). These results indicated that the EnKF optimized FNN model with higher accuracy than gradient decent algorithm. Furthermore, the KF based algorithms were able to optimize the parameters of FNN in real-time by incorporating real-time observations which is intrinsic quality of the methods. Therefore, there is no need to train the FNN models when new observations available.

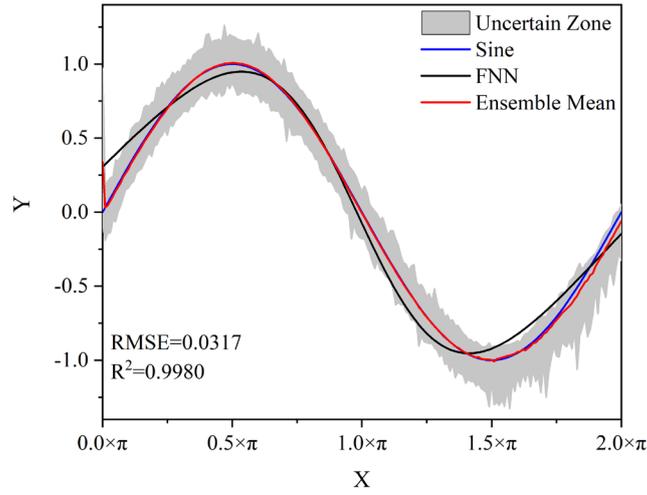

**Fig. 4.** Comparison of the results from FNN optimized by Gradient Descent (black curve) and EnKF (Red curve for Ensemble mean and grey area for uncertain zone obtained from ensembles) with Sine function.

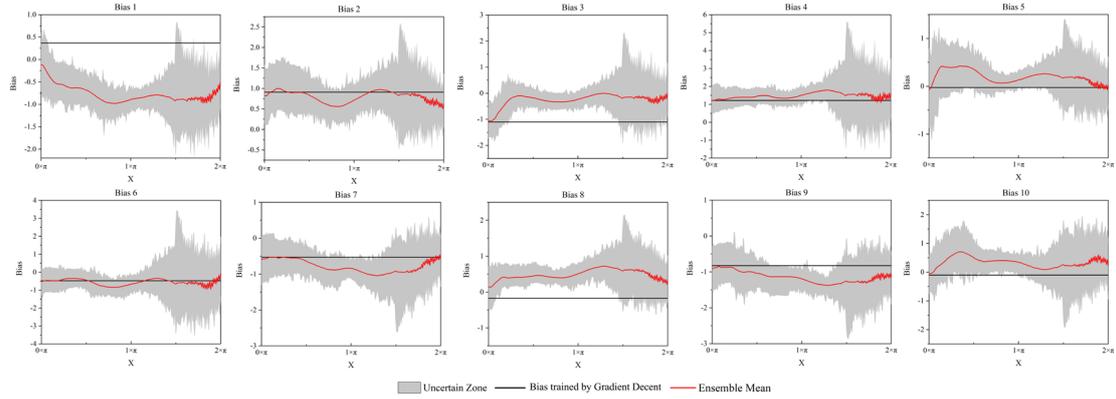

**Fig. 5.** Parameters trained by EnKF and Gradient Decent in the case of Sine function.

In the synthetic case of Mexican Hat function, the hyper-parameters of FNN and EnKF are identical with those in the Sine function case. The results calculated from FNN which optimized by different methods were shown in Fig. 6. The RMSE and $R^2$ value for FNN model optimized by gradient decent were 0.0329 and 0.9891. In the experiment of EnKF-optimized FNN, the ensemble mean of the model outputs were used to calculate the performance criteria with RMSE of 0.018 and $R^2$ of 0.9967. The better performance was attributed to the update scheme of the model states (parameters) which was shown in Eq. (7)~(11). The evolution of parameters shown in Fig. 7 indicated the update process. From Fig. 6 and Fig. 7, one can tell that the variance of parameters was larger when the difference between observations ($Y_t^0$) and simulations ($Y_t$) were large which can be explained by Eq. (7). These results indicated that EnKF was able to optimize the parameters of FNN by implementing the update process with higher accuracy.

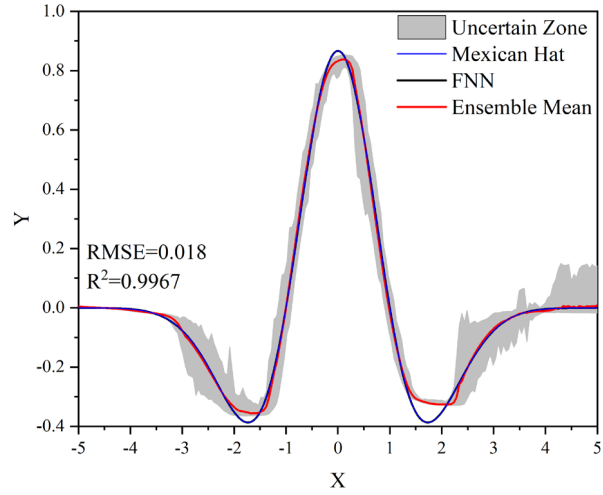

**Fig. 6.** Comparison of the results from FNN optimized by Gradient Descent (black curve) and EnKF (Red curve for Ensemble mean and grey area for uncertain zone obtained from ensembles) with Mexican Hat function.

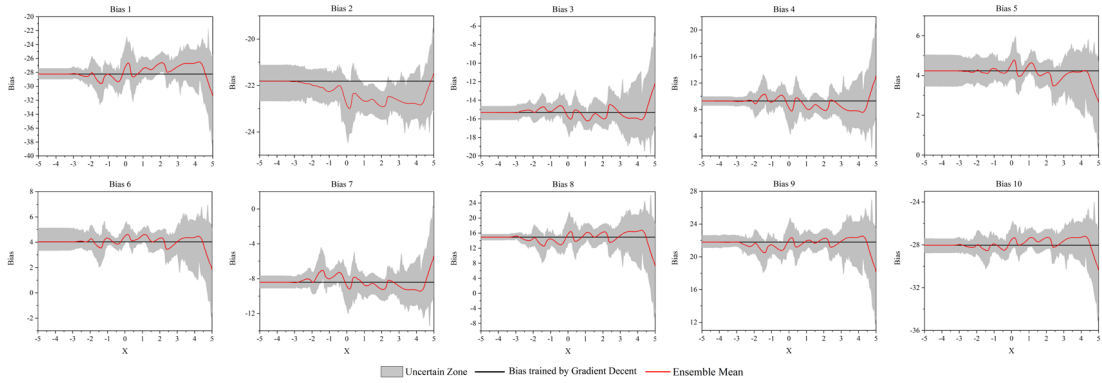

**Fig. 7.** Parameters trained by EnKF and Gradient Decent in the case of Mexican Hat function.

### 4.2. Performance of FNN model optimized by ESMDA

EnKF was used in the ESMDA to conduct the procedure of data assimilation. The FNN model was identical to the model used in the last cases. The predefined number of iterations for data assimilation $N_i$, the ensemble size $N_e$, the prior parameter covariance matrix $\varXi$ and the observation error covariance matrix $R_e$ are 3, 50, 0.1 and 0.1, respectively.

In the synthetic case of Sine function, the results of three iterations were shown in Fig. 8. In the first iteration, 50 samples of parameters were randomly generated using normal distribution with covariance matrix $\varXi$, the FNN model was executed with the generated samples to yield outputs. The uncertainty of the parameters was the largest because of the random generation. In the second iteration, the distributions of the parameters were updated by the EnKF method which significantly narrowed down the uncertain zone of the outputs. In the third iteration, the distributions of the parameters were slightly updated without significant effects on the outputs. The mean of the 50 ensembles was considered as the best estimation for the outputs in each iteration. The RMSE and $R^2$ were calculated to conduct quantitative comparisons between the observations and simulations (Table 4). Table 4 indicates better results were obtained by ESMDA than those obtained by Gradient Decent which proves the effectiveness of the ESMDA for updating the parameters. The evolution of parameters (biases from the hidden layer to the output layer) were shown in Fig. 9. The convergence of the parameters with the increase of iterations indicates the effectiveness of the ESMDA. The variances of the parameters were lower with the iterations which could be obtained from Fig. 9 by the narrowing of the uncertain zone. The mean value of the parameter in Fig. 9 which corresponds to the mean value of the trained results in Fig. 8 can be regarded as the optimal parameters for the FNN model.

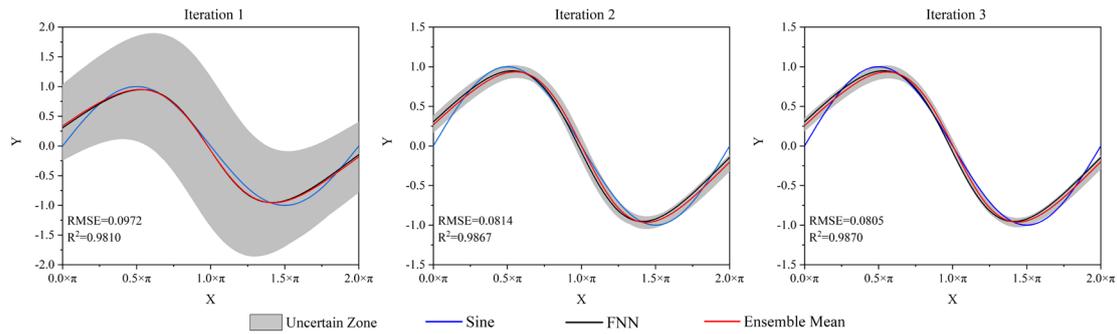

**Fig. 8.** Comparison of the results from FNN optimized by Gradient Descent (black curve) and ESMDA (Red curve for Ensemble mean and grey area for each ensemble) with Sine wave.

**Table 4.** RMSE and $R^2$ values for the Gradient Decent and ESMDA methods in the Sine function case.

|  | Gradient Decent | ESMDA | | |
| --- | --- | --- | --- | --- |
|  |  | Iteration 1 | Iteration 2 | Iteration 3 |
| RMSE | 0.0948 | 0.0972 | 0.0814 | 0.0805 |
| $R^2$ | 0.9819 | 0.9810 | 0.9867 | 0.9870 |

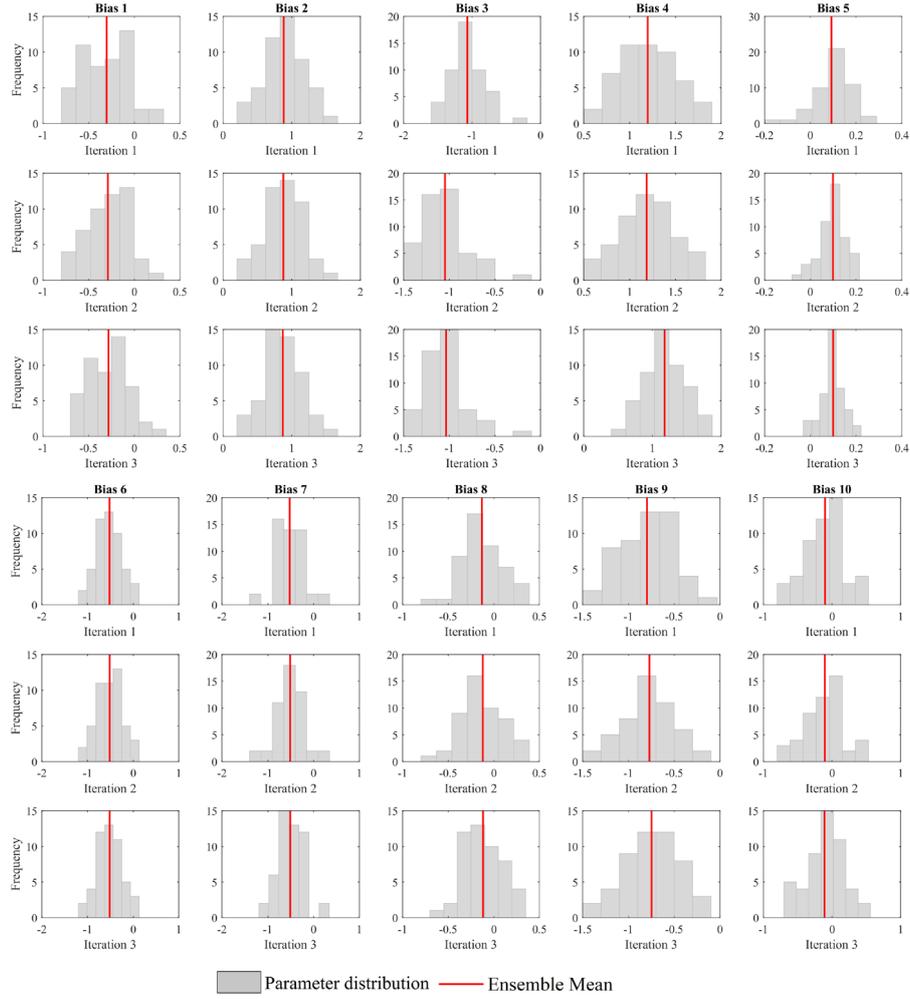

**Fig. 9.** Parameters trained by ESMDA and Gradient Decent in the Sine function case.

In the synthetic case of Mexican Hat function, the results of three iterations were shown in Fig. 10. The uncertain zone of the outputs for the first iteration was largest because of the random generation of parameters (shown in Fig. 11.). In iteration 2 and iteration 3, the uncertain zone of the outputs keep narrowing down due to the parameters updating process of ESMDA by using EnKF. It should be noted that the uncertainties of parameters were larger when the gradient of Mexican Hat function closing zero (i.e., around $x=\pm 3$, $x=\pm\sqrt{3}$ and $x=0$). The differences between the outputs of FNN and the Mexican Hat function were also relatively larger at these points. The reason may also lie in Eq. (7) as we described in Section 4.1. This phenomenon indicated the adjustment of parameters (shown in Fig. 11) according to the observations which also demonstrated the effectiveness of updating processes in ESMDA. It should also be noted that the variance of parameters was not enough to cover some points in the model outputs (i.e., $\pm\sqrt{3}$ in Fig. 10). This may be caused by the situation that only biases in the hidden layer were perturbed. Involving more parameters (for instance, weights in Eq. (1)~(2)) for perturbation and optimization may solve this problem. Quantitative comparisons between the observations and simulation were conducted by calculating RMSE and $R^2$ (Table 5.). Table 5 showed that better results were obtained by ESMDA than those obtained by Gradient Decent. The evolution of parameters was shown in Fig. 11. The variances of parameters were lower with the iterations which resulted in the narrower uncertain zones in Fig. 10.

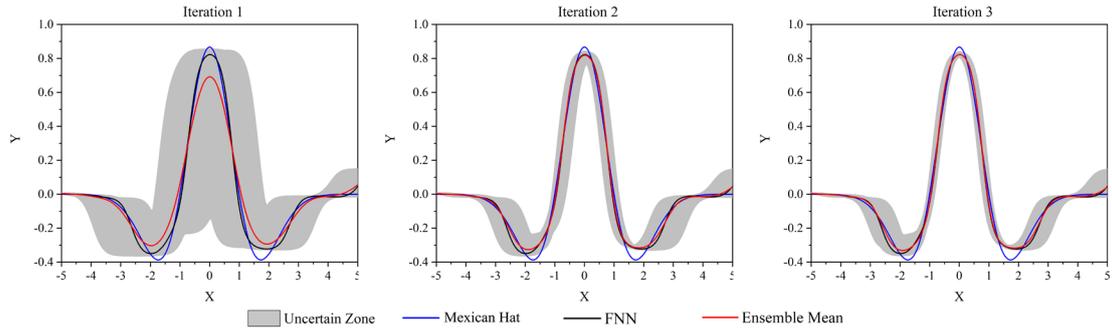

**Fig. 10.** Comparison of the results from FNN optimized by Gradient Descent (black curve) and ESMDA (Red curve for Ensemble mean and grey area for each ensemble) with Mexican Hat function.

**Table 5.** RMSE and $R^2$ values for the Gradient Decent and ESMDA methods in the Mexican Hat function case.

|  | Gradient Decent | ESMDA | | |
| --- | --- | --- | --- | --- |
|  |  | Iteration 1 | Iteration 2 | Iteration 3 |
| RMSE | 0.0329 | 0.0674 | 0.0297 | 0.0291 |
| $R^2$ | 0.9891 | 0.9544 | 0.9911 | 0.9915 |

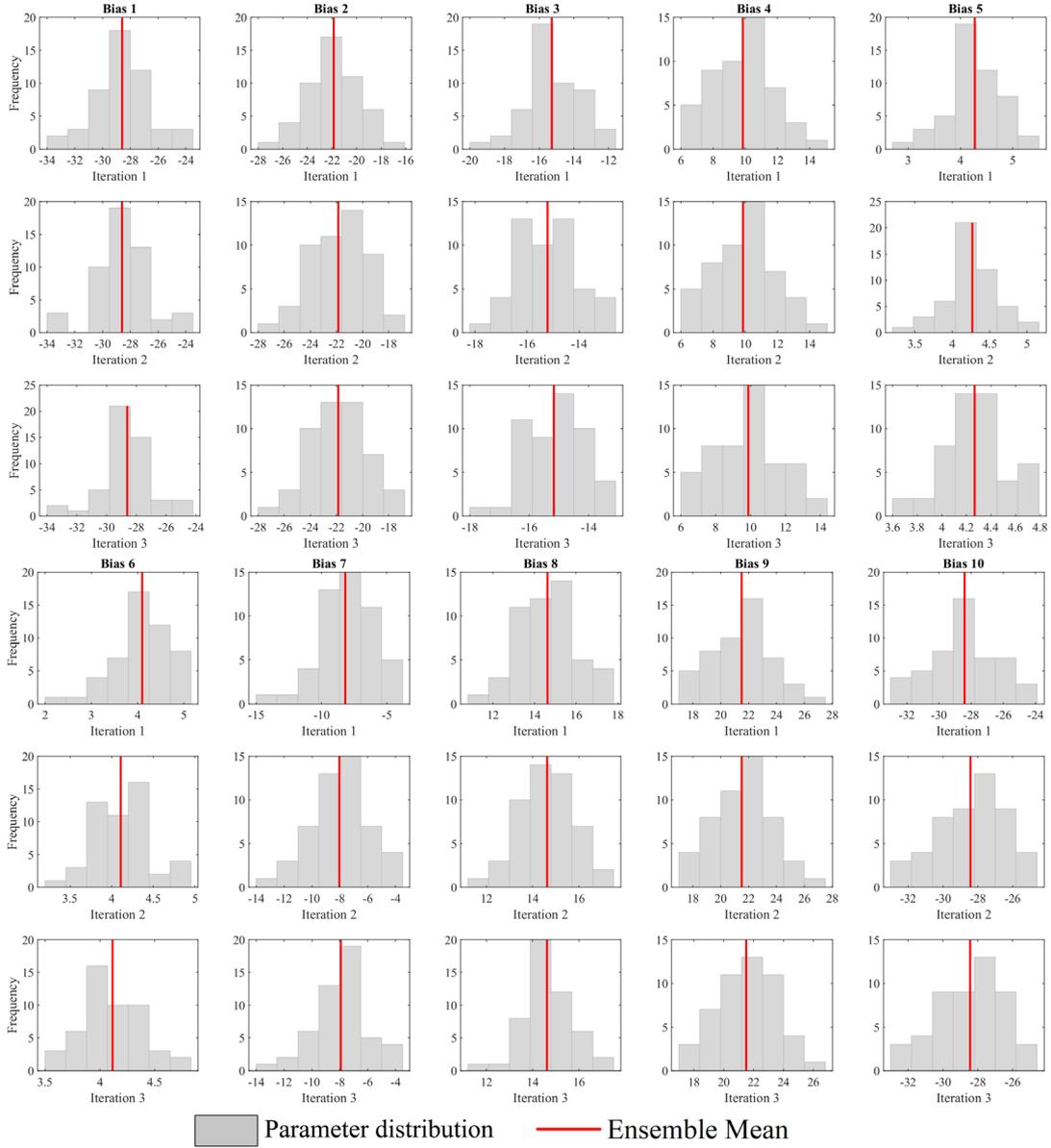

**Fig. 11.** Parameters trained by ESMDA and Gradient Decent in the Mexican Hat function case.

## 5. Conclusion

In this paper, a new training framework for neural networks based on data assimilation was proposed to avoid the calculation of gradient in the neural network training. The Feedforward Neural Networks (FNNs), Ensemble Kalman Filter (EnKF) and Ensemble Smoother with Multiple Data Assimilation (ESMDA) were used to validate the proposed framework. Synthetic cases with data generated from Sine function and Mexican Hat function was implemented to test the methods. EnKF updates the parameters when the observations available which can be regard as real-time training (online learning). ESMDA updates the parameters using all the available observations with a predefined number of iterations for data assimilation which can be regarded as normal training

(offline learning) compared to the conventional methods. The results from EnKF-optimized and ESMDA-optimized FNN model showed higher accuracy than those from gradient-decent-optimized FNN model. This indicates the effectiveness of the EnKF and ESMDA trained FNN. Furthermore, the major advantages of the proposed training methods based on the data assimilation were (1) the avoidance of calculating gradient, (2) the ability of real-time training when the observations available, (3) the uncertainty analysis for the parameters of neural networks. Although only FNN, EnKF and ESMDA were implemented as examples in this study, the potential of data assimilation algorithms on training neural networks are unlimited. Future works may include exploring new data assimilation algorithms (e.g., Particle Filter.), exploring other kinds of neural networks (e.g., Recurrent Neural Network, Graph Neural Networks), involving more parameters of neural networks, analyzing the uncertainty of parameters with data assimilation algorithms and validating the methods with real observation data.

**Acknowledgments**
The authors would like to thank the reader for their precious time and efforts and providing constructive comments (if possible) to improve the paper.